\def\year{2016}\relax
\documentclass[letterpaper]{article}
\usepackage{aaai16}
\usepackage{times}
\usepackage{helvet}
\usepackage{courier}
\usepackage{graphicx}
\usepackage{comment}
\usepackage{float}
\usepackage{amssymb}
\usepackage{color}
\usepackage{array}
\usepackage{multirow}

\newcommand\Tstrut{\rule{0pt}{2.6ex}}

\title{Transfer Learning from Deep Features for Remote Sensing and Poverty Mapping}

\author{Michael Xie \and Neal Jean \and Marshall Burke \and David Lobell \and Stefano Ermon\\
Department of Computer Science, Stanford University\\\texttt{\{xie, nealjean, ermon\}@cs.stanford.edu}\\
Department of Earth System Science, Stanford University\\
\texttt{\{mburke,dlobell\}@stanford.edu}
}

\begin{document}
\maketitle
\frenchspacing
\setlength{\belowcaptionskip}{-9pt}
\setlength{\floatsep}{-9pt}
\begin{abstract}
\begin{quote}

The lack of reliable data in developing countries is a major obstacle to sustainable development, food security, and disaster relief. Poverty data, for example, is typically scarce, sparse in coverage, and labor-intensive to obtain. Remote sensing data such as high-resolution satellite imagery, on the other hand, is becoming increasingly available and inexpensive. Unfortunately, such data is highly unstructured and currently no techniques exist to automatically extract useful insights to inform policy decisions and help direct humanitarian efforts.	
We propose a novel machine learning approach to extract large-scale socioeconomic indicators from high-resolution satellite imagery. The main challenge is that training data is very scarce, making it difficult to apply modern techniques such as Convolutional Neural Networks (CNN). We therefore propose a transfer learning approach where nighttime light intensities are used as a data-rich proxy.
We train a fully convolutional CNN model to predict nighttime lights from daytime imagery, simultaneously learning features that are useful for poverty prediction. The model learns filters identifying different terrains and man-made structures, including roads, buildings, and farmlands, without any supervision beyond nighttime lights. We demonstrate that these learned features are highly informative for poverty mapping, even approaching the predictive performance of survey data collected in the field.

\end{quote}
\end{abstract}

\section{Introduction}

New technologies fueling the Big Data revolution are creating unprecedented opportunities for designing, monitoring, and evaluating policy decisions and for directing humanitarian efforts ~\cite{abelson2014,varshney2015}. However, while rich countries are being flooded with data, developing countries are suffering from data drought.
A new data divide is emerging, with huge differences in the quantity and quality of data available. For example, some countries have not taken a census in decades, and in the past five years an estimated 230 million births have gone unrecorded~\cite{worldcounts2014}.
Even high-profile initiatives such as the Millennium Development Goals (MDGs) are affected \cite{mdg2015}. Progress based on poverty and infant mortality rate targets can be difficult to track.
Often, poverty measures must be inferred from small-scale and expensive household surveys, effectively rendering many of the poorest people invisible.

Remote sensing, particularly satellite imagery, is perhaps the only cost-effective technology able to provide data at a global scale. Within ten years, commercial services are expected to provide sub-meter resolution images everywhere at a fraction of current costs \cite{murthy2014}. This level of temporal and spatial resolution could provide a wealth of data towards sustainable development. Unfortunately, this raw data is also highly unstructured, making it difficult to extract actionable insights at scale.

In this paper, we propose a machine learning approach for extracting socioeconomic indicators from raw satellite imagery. In the past five years, deep learning approaches applied to large-scale datasets such as ImageNet have revolutionized the field of computer vision, leading to dramatic improvements in fundamental tasks such as object recognition~\cite{russakovsky2014}. However, the use of contemporary techniques for the analysis of remote sensing imagery is still largely unexplored. Modern approaches such as Convolutional Neural Networks (CNN) can, in principle, be directly applied to extract socioeconomic factors, but the primary challenge is a lack of training data. While such data is readily available in the United States and other developed nations, it is extremely scarce in Africa where these techniques would be most useful.

We overcome this lack of training data by using a sequence of transfer learning steps and a convolutional neural network model. The idea is to leverage available datasets such as ImageNet to extract features and high-level representations that are useful for the task of interest, i.e., extracting socioeconomic data for poverty mapping. Similar strategies have proven quite successful in the past. For example, image features from the \texttt{Overfeat} network trained on ImageNet for object classification achieved state-of-the-art results on tasks such as fine-grained recognition, image retrieval, and attribute detection~\cite{razavian2014}.

Pre-training on ImageNet is useful for learning low-level features such as edges. 
However, ImageNet consists only of object-centric images, while satellite imagery is captured from an aerial, bird's-eye view. We therefore employ a second transfer learning step, where nighttime light intensities are used as a proxy for economic activity.
Specifically, we start with a CNN model pre-trained for object classification on ImageNet and learn a modified network that predicts nighttime light intensities from daytime imagery. To address the trade-off between fixed image size and information loss from image scaling, we use a fully convolutional model that takes advantage of the full satellite image. We show that transfer learning succeeds in learning features relevant not only for nighttime light prediction but also for poverty mapping. For instance, the model learns filters identifying man-made structures such as roads, urban areas, and fields without any supervision beyond nighttime lights, i.e., without any labeled examples of roads or urban areas (Figure \ref{fig:filter1}). We demonstrate that these features are highly informative for poverty mapping and capable of approaching the predictive performance of survey data collected in the field.

\section{Problem Setup}

We begin by reviewing transfer learning and convolutional neural networks, the building blocks of our approach.

\subsection{Transfer Learning}

We formalize transfer learning as in \cite{pan2010}: A \emph{domain} $\mathcal{D}=\{\mathcal{X}, P(\mathcal{X})\}$ consists of a feature space $\mathcal{X}$ and a marginal probability distribution $P(\mathcal{X})$. Given a domain, a \emph{task} $\mathcal{T}=\{\mathcal{Y}, f(\cdot)\}$ consists of a label space $\mathcal{Y}$ and a predictive function $f(\cdot)$ which models $P(y \vert x)$ for $y \in \mathcal{Y}$ and $x \in \mathcal{X}$. 
Given a source domain $\mathcal{D}_S$
and learning task $\mathcal{T}_S$, and a target domain $\mathcal{D}_T$ and learning task
$\mathcal{T}_T$, \emph{transfer learning} aims to improve the learning of the
target predictive function $f_T(\cdot)$ in $\mathcal{T}_T$ using the knowledge from
$\mathcal{D}_S$ and $\mathcal{T}_S$, where $\mathcal{D}_S \neq \mathcal{D}_T$, $\mathcal{T}_S \neq \mathcal{T}_T$, or both.
Transfer learning is particularly relevant when, given labeled \textit{source domain data} 
$D_S$
and \textit{target domain data} 
$D_T$, we find that $|D_T| \ll |D_S| $.

In our setting, we are interested in more than two related learning tasks. We generalize the formalism by representing the multiple source-target relationships as a \emph{transfer learning graph}. First, we define a \textit{transfer learning problem} $\mathcal{P}=(\mathcal{D}, \mathcal{T})$ as a domain-task pair. The \emph{transfer learning graph} is then defined as follows:
A transfer learning graph $G=(\mathcal{V},\mathcal{E})$ is a directed acyclic graph where vertices $\mathcal{V}=\{\mathcal{P}_1, \cdots, \mathcal{P}_v\}$ are transfer learning problems and $\mathcal{E} = \{(\mathcal{P}_{i_1},\mathcal{P}_{j_1}), \cdots, (\mathcal{P}_{i_e}, \mathcal{P}_{j_e})\}$ is an edge set. For each transfer learning problem $\mathcal{P}_i=(\mathcal{D}_i, \mathcal{T}_i) \in \mathcal{V}$, the aim is to improve the learning of the
target predictive function $f_i(\cdot)$ in $\mathcal{T}_i$ using the knowledge in
$\cup_{(j,i) \in \mathcal{E}} \mathcal{P}_j$.

\subsection{Convolutional Neural Networks}
Deep learning approaches are based on automatically learning nested, hierarchical representations of data. Deep feed-forward neural networks are the typical example of deep learning models. Convolutional Neural Networks (CNN) include convolutional operations over the input and are designed specifically for vision tasks.
Convolutional filters are useful for encoding translation invariance, a key concept for discovering useful features in images~\cite{bouvrie2006}. 

A CNN is a general function approximator defined by a set of convolutional and fully connected layers ordered such that the output of one layer is the input of the next. For image data, the first layers of the network typically learn low-level features such as edges and corners, and further layers learn high-level features such as textures and objects~\cite{zeiler2013}.
Taken as a whole, a CNN is a mapping from tensors to feature vectors, which become the input for a final classifier.
A typical convolutional layer maps a tensor $x \in \mathbb{R}^{h \times w \times d}$ to $g_i \in \mathbb{R}^{\hat{h} \times \hat{w} \times \hat{d}}$ such that
$$g_i = p_i(f_i(W_i \ast x + b_i)),$$ 
where for the $i$-th convolutional layer, $W_i \in \mathbb{R}^{l \times l \times  \hat{d}}$ is a tensor of $\hat{d}$ convolutional filter weights of size $l \times l$, $(\ast)$ is the 2-dimensional convolution operator over the last two dimensions of the inputs, $b_i$ is a bias term, $f_i$ is an element-wise nonlinearity function (e.g., a rectified linear unit or ReLU), and $p_i$ is a pooling function.
The output dimensions $\hat{h}$ and $\hat{w}$ depend on the stride and zero-padding parameters of the layer, which control how the convolutional filters slide across the input.
For the first convolutional layer, the input dimensions $h$, $w$, and $d$ can be interpreted as height, width, and number of color channels of an input image, respectively.  

In addition to convolutional layers, most CNN models have fully connected layers in the final layers of the network. Fully connected layers map an unrolled version of the input $\hat{x} \in \mathbb{R}^{hwd}$, which is a one-dimensional vector of the elements of a tensor $x \in \mathbb{R}^{h \times w \times d}$, to an output $g_i\in \mathbb{R}^k$ such that
$$g_i=f_i(W_i\hat{x}+b_i),$$
where $W_i \in \mathbb{R}^{k\times hwd}$ is a weight matrix, $b_i$ is a bias term, and $f_i$ is typically a ReLU nonlinearity function. 
The fully connected layers encode the input examples as feature vectors, which are used as inputs to a final classifier. Since the fully connected layer looks at the entire input at once, these feature vectors ``summarize'' the input into a feature vector for classification.
The model is trained end-to-end using minibatch gradient descent and backpropagation.

After training, the output of the final fully connected layer can be interpreted as an encoding of the input as a feature vector that facilitates classification. These features often represent complex compositions of the lower-level features extracted by the previous layers (e.g., edges and corners) and can range from grid patterns to animal faces~\cite{zeiler2013,le2012}.

\subsection{Combining Transfer Learning and Deep Learning}

The low-level and high-level features learned by a CNN on a source domain can often be transferred to augment learning in a different but related target domain. For target problems with abundant data, we can transfer low-level features, such as edges and corners,
and learn new high-level features specific to the target problem.
For target problems with limited amounts of data, learning new high-level features is difficult. However, if the source and target domain are sufficiently similar, the feature representation learned by the CNN on the source task can also be used for the target problem. Deep features extracted from CNNs trained on large annotated datasets of images have been used as generic features very effectively for a wide range of vision tasks ~\cite{donahue2013,oquab2014}.

\section{Transfer Learning for Poverty Mapping}

In our approach to poverty mapping using satellite imagery, we construct a linear chain transfer learning graph with $\mathcal{V}=\{\mathcal{P}_1, \mathcal{P}_2, \mathcal{P}_3\}$ and $\mathcal{E}=\{(\mathcal{P}_1, \mathcal{P}_2), (\mathcal{P}_2, \mathcal{P}_3)\}$. The first transfer learning problem $\mathcal{P}_1$ is object recognition on ImageNet~\cite{russakovsky2014}; the second problem $\mathcal{P}_2$ is predicting nighttime light intensity from daytime satellite imagery; the third problem $\mathcal{P}_3$ is predicting poverty from daytime satellite imagery. Recognizing the differences between ImageNet data and satellite imagery, we use the intermediate problem $\mathcal{P}_2$ to learn about the bird's-eye viewpoint of satellite imagery and extract features relevant to socioeconomic development.

\subsection{ImageNet to Nighttime Lights}

ImageNet is an object classification image dataset of over 14 million images with 1000 class labels that, along with CNN models, have fueled major breakthroughs in many vision tasks~\cite{russakovsky2014}. CNN models trained on the ImageNet dataset are recognized as good generic feature extractors, with low-level and mid-level features such as edges and corners that are able to generalize to many new tasks~\cite{donahue2013,oquab2014}. Our goal is to transfer knowledge from the ImageNet object recognition challenge ($\mathcal{P}_1$) to the target problem of predicting nighttime light intensity from daytime satellite imagery ($\mathcal{P}_2$).

In $\mathcal{P}_1$, we have an object classification problem with source domain data $D_1=\{({x_1}_i, {y_1}_i)\}$ from ImageNet that consists of natural images ${x_1}_i \in \mathcal{X}_1$ and object class labels. 
In $\mathcal{P}_2$, we have a nighttime light intensity prediction problem with target domain data $D_2=\{({x_2}_i, {y_2}_i\})$ that consists of daytime satellite images ${x_2}_i \in \mathcal{X}_2$ and nighttime light intensity labels. 
Although satellite data is still in the space of image data, satellite imagery presents information from a bird's-eye view and at a much different scale than the object-centric ImageNet dataset ($P(\mathcal{X}_1) \neq P(\mathcal{X}_2)$). Previous work in domains with images fundamentally different from normal ``human-eye view'' images typically resort to curating a new, specific dataset such as Places205 \cite{zhou2014}. In contrast, our transfer learning approach does not require human annotation and is much more scalable. Additionally, unsupervised approaches such as autoencoders may waste representational capacity on irrelevant features, while the nighttime light labels guide learning towards features relevant to wealth and economic development.

The National Oceanic and Atmospheric Administration (NOAA) provides annual nighttime images of the world with 30 arc-second resolution, or about 1 square kilometer \cite{noaa2014}. The light intensity values are averaged and denoised for each year to ensure that ephemeral light sources do not affect the data.

\begin{figure}
\centering
  \includegraphics[width=0.8\linewidth]{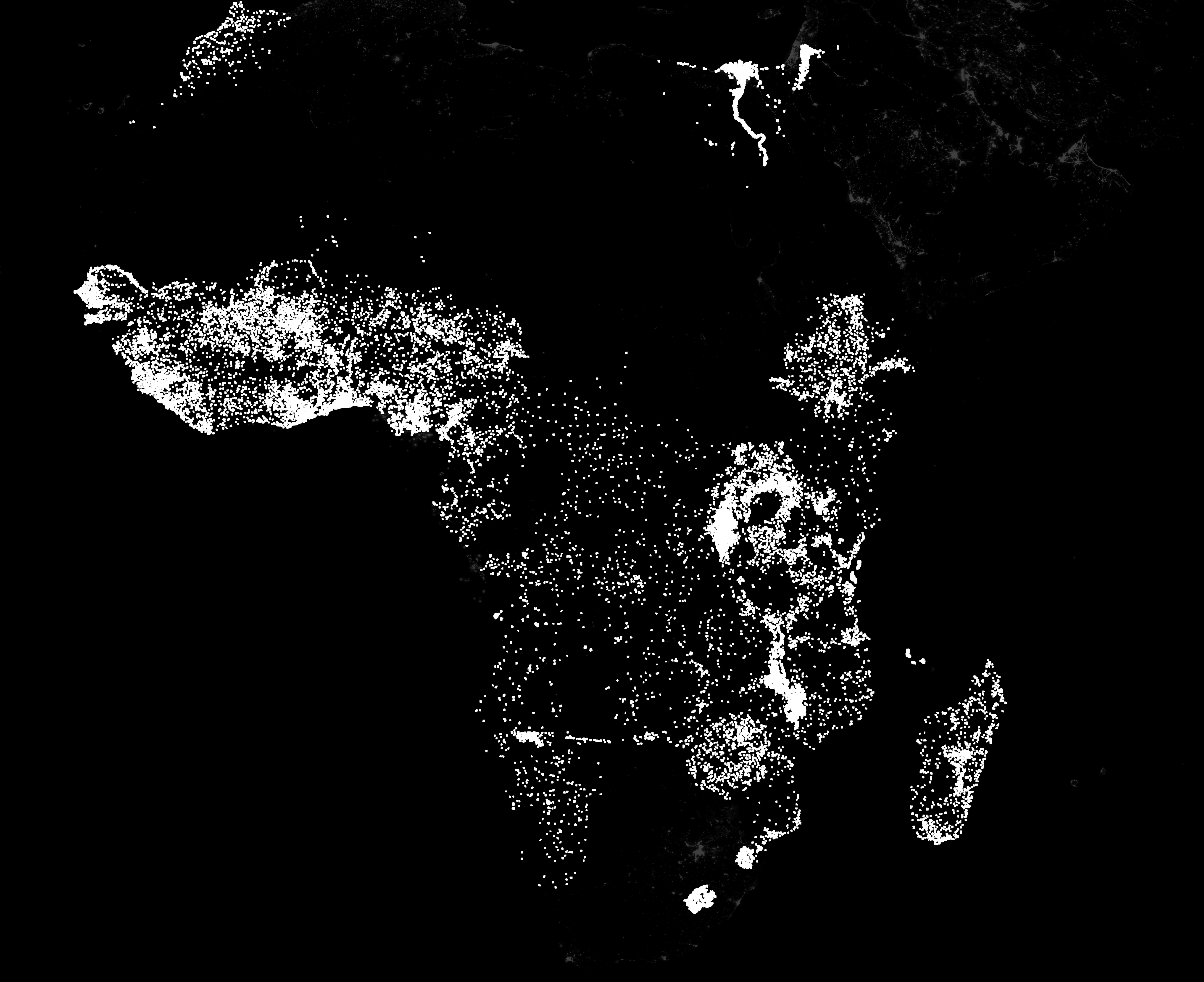}
  \caption{Locations (in white) of 330,000 sampled daytime images near DHS survey locations for the nighttime light intensity prediction problem.}
  \label{fig:sampled}
\end{figure}

The nighttime light dataset $D_2$ is constructed as follows: The Demographic Health Survey (DHS) Program conducts nationally representative surveys in Africa that focus mainly on health outcomes~\cite{dhs2015}. Predicting health outcomes is beyond the scope of this paper; however, the DHS surveys offer the most comprehensive data available for Africa. Thus, we use DHS survey locations as guidelines for sampling training images (see Figure \ref{fig:sampled}). 
Images in $D_2$ are daytime satellite images randomly sampled near DHS survey locations in Africa.
Satellite images are downloaded using the Google Static Maps API, each with $400 \times 400$ pixels at zoom level 16, resulting in images similar in size to pixels in the NOAA nighttime lights data. The aggregate dataset $D_2$ consists of over 330,000 images, each labeled with an integer nighttime light intensity value ranging from 0 to 63\footnote{Nighttime light intensities are from 2013, while the daytime satellite images are from 2015. We assume that the areas under study have not changed significantly in this two-year period, but this temporal mismatch is a potential source of error.}. We further subsample and bin the data using a Gaussian mixture model, as detailed in the companion technical report~\cite{DBLP:journals/corr/XieJBLE15}.

\subsection{Nighttime Lights to Poverty Estimation}

The final and most important learning task $\mathcal{P}_3$ is that of predicting poverty from satellite imagery, for which we have very limited training data. Our goal is to transfer knowledge from $\mathcal{P}_2$, a data-rich problem, to $\mathcal{P}_3$.

The target domain data $D_3=\{({x_3}_i, {y_3}_i)\}$ consists of satellite images ${x_3}_i \in \mathcal{X}_3$ from the feature space of satellite images of Uganda and a \textbf{limited number} of poverty labels ${y_3}_i \in \mathcal{Y}_3$, detailed below. The source data is $D_2$, the nighttime lights data. 
Here, the input feature space of images is similar in both the source and target domains, drawn from a similar distribution of images (satellite images) from related areas (Africa and Uganda), 
implying that $\mathcal{X}_2 =\mathcal{X}_3$, $P(\mathcal{X}_2) \approx P(\mathcal{X}_3)$. The source (lights) and target (poverty) tasks both have economic elements, but are quite different.

The poverty training data $D_3$ relies on the Living Standards Measurement Study (LSMS) survey conducted in Uganda by the Uganda Bureau of Statistics between 2011 and 2012 \cite{lsms2012}. 
The LSMS survey consists of data from 2,716 households in Uganda, which are grouped into 643 unique location groups.
The average latitude and longitude location of the households within each group is given, with added noise of up to 5km in each direction. Individual household locations are withheld to preserve anonymity.
In addition, each household has a binary poverty label based on expenditure data from the survey.
We use the majority poverty classification of households in each group as the overall location group poverty label.
For a given group, we sample approximately 100 1km$\times$1km images tiling a 10km $\times$ 10km area centered at the average household location as input.
This defines the probability distribution $P(\mathcal{X}_3)$ of the input images for the poverty classification problem $\mathcal{P}_3$.

\section{Predicting Nighttime Light Intensity}
Our first goal is to transfer knowledge from the ImageNet object recognition task to the nighttime light intensity prediction problem.
We start with a CNN model with parameters trained on ImageNet, then modify the network to adapt it to the new task (i.e., change the classifier on the last layer to reflect the new nighttime light prediction task). We train on the new task using SGD with momentum, using ImageNet parameters as initialization to achieve knowledge transfer.

We choose the  \texttt{VGG F} model trained on ImageNet as the starting CNN model ~\cite{chatfield2014}. The \texttt{VGG F} model has 8 convolutional and fully connected layers.
Like many other ImageNet models, the \texttt{VGG F} model accepts a fixed input image size of $224 \times 224$ pixels. Input images in $D_2$, however, are $400 \times 400$ pixels, corresponding to the resolution of the nighttime lights data.

We consider two ways of adapting the original \texttt{VGG F} network. The first approach is to keep the structure of the network (except for the final classifier) and crop the input to $224 \times 224$ pixels (\textbf{random cropping}). This is a reasonable approach, as the original model was trained by cropping $224 \times 224$ images from a larger $256 \times 256$ image~\cite{chatfield2014}. 
Ideally, we would evaluate the network at multiple crops of the $400\times400$ input and average the predictions to leverage the context of the entire input image. However, doing this explicitly with one forward pass for each crop would be too costly. Alternatively, if we allow the multiple crops of the image to overlap, we can use a convolution to compute scores for each crop simultaneously, gaining speed by reusing filter outputs at all layers. We therefore propose a fully convolutional architecture (\textbf{fully convolutional}).

\begin{figure*}
\centering
  \includegraphics[width=\linewidth]{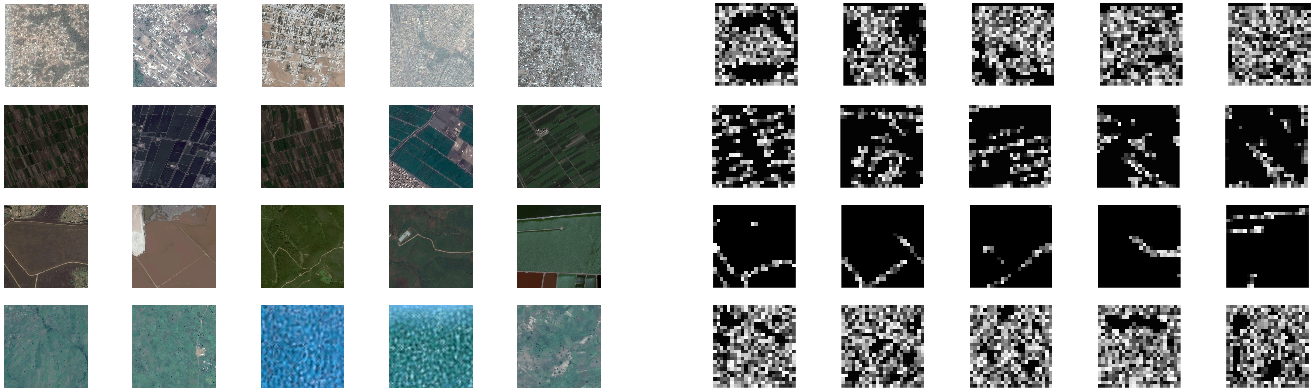}
  \caption{\textbf{Left:} Each row shows five maximally activating images for a different filter in the fifth convolutional layer of the CNN trained on the nighttime light intensity prediction problem. The first filter (first row) activates for urban areas. The second filter activates for farmland and grid-like patterns. The third filter activates for roads. The fourth filter activates for water, plains, and forests, terrains contributing similarly to nighttime light intensity. The only supervision used is nighttime light intensity, i.e., no labeled examples of roads or farmlands are provided. \textbf{Right:} Filter activations for the corresponding images on the left. Filters mostly activate on the relevant portions of the image. For example, in the third row, the strongest activations coincide with the road segments. Best seen in color. See the companion technical report for more visualizations~\cite{DBLP:journals/corr/XieJBLE15}. Images from Google Static Maps.}
  \label{fig:filter1}
\end{figure*}

\subsection{Fully Convolutional Model}

Fully convolutional models have been used successfully for spatial analysis of arbitrary size inputs~\cite{wolf1994,long2014}. 
We construct the fully convolutional model by converting the fully connected layers of the \texttt{VGG F} network to convolutional layers. This allows the network to efficiently ``slide'' across a larger input image and make multiple evaluations of different parts of the image, incorporating all available contextual information.

Given an unrolled $h \times w \times d$-dimensional input $x \in \mathbb{R}^{hwd}$, fully connected layers perform a matrix-vector product
$$\hat{x}=f(Wx+b)$$
where $W \in \mathbb{R}^{k\times hwd}$ is a weight matrix, $b$ is a bias term, $f$ is a nonlinearity function, and $\hat{x} \in \mathbb{R}^k$ is the output. In the fully connected layer, we take $k$ inner products with the unrolled $x$ vector. Thus, given a differently sized input, it is unclear how to evaluate the dot products.

We replace a fully connected layer by a convolutional layer with $k$ convolutional filters of size $h\times w$, the same size as the input. The filter weights are shared across all channels, which means that the convolutional layer actually uses fewer parameters than the fully connected layer. Since the filter size is matched with the input size, we can take an element-wise product and add, which is equivalent to an inner product. This results in a scalar output for each filter, creating an output $\hat{x} \in \mathbb{R}^{1 \times 1 \times k}$.
Further fully connected layers are converted to convolutional layers with filter size $1\times1$, matching the new input $\hat{x} \in \mathbb{R}^{1 \times 1 \times k}$. Fully connected layers are usually the last layers of the network, while all previous layers are typically convolutional. After converting fully connected layers to convolutional layers, the entire network becomes convolutional, allowing the outputs of each layer to be reused as the convolution slides the network over a larger input.
Instead of a scalar output, the new output is a 2-dimensional map of filter activations.

In our fully convolutional model, the $400\times400$ input produces an output of size $2\times2\times4096$, which represents the scores of four (overlapping) quadrants of the image for 4096 features. The regional scores are then averaged to obtain a 4096-dimensional feature vector that becomes the final input to the classifier predicting nighttime light intensity.

\subsection{Training and Performance Evaluation}

Both CNN models are trained using minibatched gradient descent with momentum. 
Random mirroring is used for data augmentation, along with 50\% dropout on convolutional layers replacing fully connected layers. The learning rate begins at 1e-6, a hundredth of the ending learning rate of the \texttt{VGG} model. All other hyperparameters are the same as in the \texttt{VGG} model as described in \cite{chatfield2014}. The \texttt{VGG} model parameters are obtained from the Caffe Model Zoo, and all networks are trained with Caffe \cite{jia2014}. 
The fully convolutional model is fine-tuned from the pre-trained parameters of the \texttt{VGG F} model, but it randomly initializes the convolutional layers that replace fully connected layers.
 
In the process of cropping, the random cropping model throws away over 68\% of the input image when predicting the class scores, losing much of the spatial context. The random cropping model achieved a validation accuracy of 70.04\% after 400,200 SGD iterations.
In comparison, the fully convolutional model achieved 71.58\% validation accuracy after only 223,500 iterations. Both models were trained in roughly three days. Despite reinitializing the final convolutional layers from scratch, the fully convolutional model exhibits faster learning and better performance. The final fully convolutional model achieves a validation accuracy of 71.71\%, trained over 345,000 iterations.

\section{Visualizing the Extracted Features}

Nighttime lights are used as a data-rich proxy, so absolute performance on this task is not directly relevant for poverty mapping. The goal is to learn high-level features that are indicative of economic development and can be used for poverty mapping in the spirit of transfer learning.

We visualize the filters learned by the fully convolutional network by inspecting the 25 maximally activating images for each filter (Figure~\ref{fig:filter1}, left and the companion technical report for more visualizations~\cite{DBLP:journals/corr/XieJBLE15}). Activation levels for filters in the middle of the network are obtained by passing the images forward through the filter, applying the ReLU nonlinearity, and then averaging the map of activation values.
We find that many filters learn to identify semantically meaningful features such as urban areas, water, roads, barren land, forests, and farmland. Amazingly, \textbf{these features are learned without direct supervision}, in contrast to previous efforts to extract features from aerial imagery, which have relied heavily on large amounts of expert-labeled data, e.g., labeled examples of roads ~\cite{mnih2010,mnih2012}. To confirm the semantics of the filters,  we visualize their activations for the same set of images (Figure~\ref{fig:filter1}, right). These maps confirm our interpretation by identifying the image parts that are most responsible for activating the filter. For example, the filter in the third row mostly activates on road segments.
These features are extremely useful socioeconomic indicators and suggest that transfer learning to the poverty task is possible.

\begin{figure*}[tbp]
\centering
  \includegraphics[width=\linewidth]{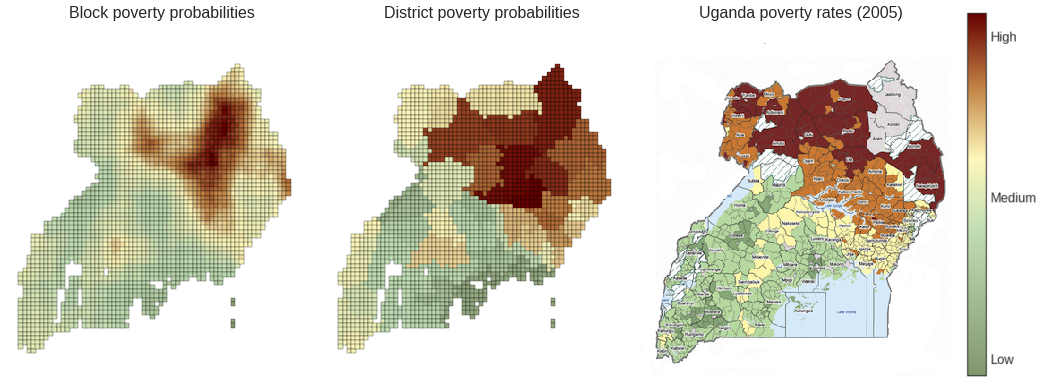}
  \caption{\textbf{Left:} Predicted poverty probabilities at a fine-grained 10km $\times$ 10km block level. \textbf{Middle:} Predicted poverty probabilities aggregated at the district-level. \textbf{Right:} 2005 survey results for comparison~\cite{wri2009}.
  }
  \label{fig:uganda_map}
\end{figure*}
\section{Poverty Estimation and Mapping}

The first target task we consider is to predict whether the majority of households are above or below the poverty threshold for 643 groups of households in Uganda. 

Given the limited amount of training data, we do not attempt to learn new feature representations for the target task. Instead, we directly use the feature representation learned by the CNN on the nighttime lights task ($\mathcal{P}_2$). Specifically, we evaluate the CNN model on new input images and feed the feature vector produced in the last layer as input to a logistic regression classifier, which is trained on the poverty task (\textbf{transfer model}). Approximately 100 images in a 10km $\times$ 10km area around the average household location of each group are used as input. We compare against the performance of a classifier with features from the \texttt{VGG F} model trained on ImageNet only (\textbf{ImageNet model}), i.e., without transfer learning from nighttime lights. In both the ImageNet model and the transfer model, the feature vectors are averaged over the input images for each group.

The Uganda LSMS survey also includes household-specific data. We extract the features that could feasibly be detected with remote sensing techniques, including roof material, number of rooms, house type, distances to various infrastructure points, urban or rural classification, annual temperature, and annual precipitation. These survey features are then averaged over each household group. The performance of the classifier trained with survey features (\textbf{survey model}) represents the gold standard for remote sensing techniques. 
We also compare with a classifier trained using the nighttime light intensities themselves as features (\textbf{lights model}).
The nighttime light features consist of the average light intensity, summary statistics, and histogram-based features for each area.
Finally, we compare with a classifier trained using a concatenation of ImageNet features and nighttime light features (\textbf{ImageNet + lights model}), an explicit way of combining information from both source problems. 

All models are trained using a logistic regression classifier with L1 regularization using a nested 10-fold cross validation (CV) scheme, 
where the inner CV is used to tune a new regularization parameter for each outer CV iteration.
The regularization parameter is found by a two-stage approach: a coarse linearly spaced search is followed by a finer linearly spaced search around the best value found in the coarse search. The tuned regularization parameter is then validated on the test set of the outer CV loop, which remained unseen as the parameter was tuned. All performance metrics are averaged over the outer 10 folds and reported in Table \ref{fig:table1}.

Our transfer model significantly outperforms every model except the survey model in every measure except recall. Notably, the transfer model outperforms all combinations of features from the source problems, implying that transfer learning was successful in learning novel and useful features. Remarkably, our transfer model based on remotely sensed data approaches the performance of the survey model based on data expensively collected in the field.  
As a sanity check, we find that using simple traditional computer vision features such as HOG and color histograms only achieves slightly better performance than random guessing. This further affirms that the transfer learning features are nontrivial and contain information more complex than just edges and colors.
\begin{table}
    \begin{center}
    \scalebox{0.97}{\begin{tabular}{|c|ccccc|}
    \hline
     &\multirow{2}{*}{{\small Survey}}&\multirow{2}{*}{{\small ImgNet}}&\multirow{2}{*}{{\small Lights}}&{{\small ImgNet}}&\multirow{2}{*}{{\small Transfer}}\\ 
     & & & & {\small +Lights}&\\
     \hline
    {\small Accuracy} \Tstrut & 0.754 & 0.686 & 0.526 & 0.683 & \textbf{0.716} \\ 
    {\small F1 Score}  & 0.552 & 0.398 & 0.448 & 0.400 & \textbf{0.489}\\  
    {\small Precision} & 0.450 & 0.340 & 0.298 &  0.338 & \textbf{0.394}\\ 
    {\small Recall} & 0.722 & 0.492 & 0.914 & 0.506 & 0.658 \\ 
    {\small AUC} & 0.776 & 0.690 & 0.719 & 0.700 & \textbf{0.761} \\ 
    \hline
    \end{tabular}}
    \caption{Cross validation test performance for predicting aggregate-level poverty measures. Survey is trained on survey data collected in the field. All other models are based on satellite imagery. Our transfer learning approach outperforms all non-survey classifiers significantly in every measure except recall, and approaches the survey model.}
    \label{fig:table1}
    \end{center}
\end{table}

To understand the high recall of the lights model, we analyze the conditional probability of predicting ``poverty'' given that the average light intensity is zero: The lights model predicts ``poverty'' almost 100\% of the time, though only 51\% of groups with zero average intensity are actually below the poverty line. 
Furthermore, only 6\% of groups with nonzero average light intensity are below the poverty line, explaining the high recall of the lights model.
In contrast, the transfer model predicts ``poverty'' in 52\% of groups where the average nighttime light intensity is 0, more accurately reflecting the actual probability.
The transfer model features (visualized in Figure \ref{fig:filter1}) clearly contain additional, meaningful information beyond what nighttime lights can provide. The fact that the transfer model outperforms the lights model indicates that transfer learning has succeeded.

\subsection{Mapping Poverty Distribution}

Using our transfer model, we can scalably and inexpensively construct fine-grained poverty maps at the country or even continent level. We evaluate this capability by estimating a country-level poverty map for Uganda. We download over 370,000 satellite images covering Uganda and estimate poverty probabilities at 1km $\times$ 1km resolution with the transfer model. Areas where the model assigns a low probability of being impoverished are colored green, while areas assigned a high risk of poverty are colored red.
A 10km $\times$ 10km resolution map is shown in Figure \ref{fig:uganda_map} (left), smoothed at a 0.5 degree radius for easy identification of dominant spatial patterns. Notably, poverty reduction in northern Uganda is lagging ~\cite{poverty2014}.
Figure \ref{fig:uganda_map} (middle) shows poverty estimates aggregated at the district level.
As a validity check, we qualitatively compare this map against the most recent map of poverty rates available (Figure \ref{fig:uganda_map}, right), which is based on 2005 survey data~\cite{wri2009}. 
This data is now a decade old, but it loosely corroborates the major patterns in our predicted distribution. Whereas current maps are coarse and outdated, our method offers much finer temporal and spatial resolution and an inexpensive way to evaluate poverty at a global scale.

\section{Conclusion}

We introduce a new transfer learning approach for analyzing satellite imagery that leverages recent deep learning advances and multiple data-rich proxy tasks to learn high-level feature representations of satellite images. This knowledge is then transferred to data-poor tasks of interest in the spirit of transfer learning. We demonstrate an application of this idea in the context of poverty mapping and introduce a fully convolutional CNN model that, without explicit supervision, learns to identify complex features such as roads, urban areas, and various terrains. Using these features, we are able to approach the performance of data collected in the field for poverty estimation. Remarkably, our approach outperforms models based directly on the data-rich proxies used in our transfer learning pipeline. Our approach can easily be generalized to other remote sensing tasks and has great potential to help solve global sustainability challenges.

\section{Acknowledgements}
We acknowledge the support of the Department of Defense through the National Defense Science and Engineering Graduate Fellowship Program. We would also like to thank NVIDIA Corporation for their contribution to this project through an NVIDIA Academic Hardware Grant.

\bibliography{egbib}
\bibliographystyle{aaai}

\clearpage
\newpage
\appendix

\section*{Appendix: Data Preparation}
Of the over 330,000 images in $D_2$, 58\% of the images are labeled with zero nighttime light intensity.  
This is a very unbalanced dataset, which is difficult to learn from. We thus alter $P(\mathcal{X}_2)$ by choosing to upsample images with higher intensities and downsample images with zero intensity until the least frequently occurring intensity has at least half the occurrences of the most frequent class. Observing that examples at similar intensity levels are hard to distinguish, a relabeled dataset with three integer label bins ranging from 0 to 2 was created by clustering the intensity levels by frequency using a 3-component Gaussian mixture model and then balancing the dataset for three classes. The 3-class balanced dataset consists of 150,000 training images and 8,000 validation images. We find that binning by frequency separates the intensity classes more intuitively. Because predicting absolute nighttime light intensity is not the final target task, it is more important for the model to extract features that are semantically meaningful than to predict light intensity with the highest accuracy.
\section*{Filter Visualizations}
\begin{figure*}[b!]
\centering
  \includegraphics[width=\linewidth]{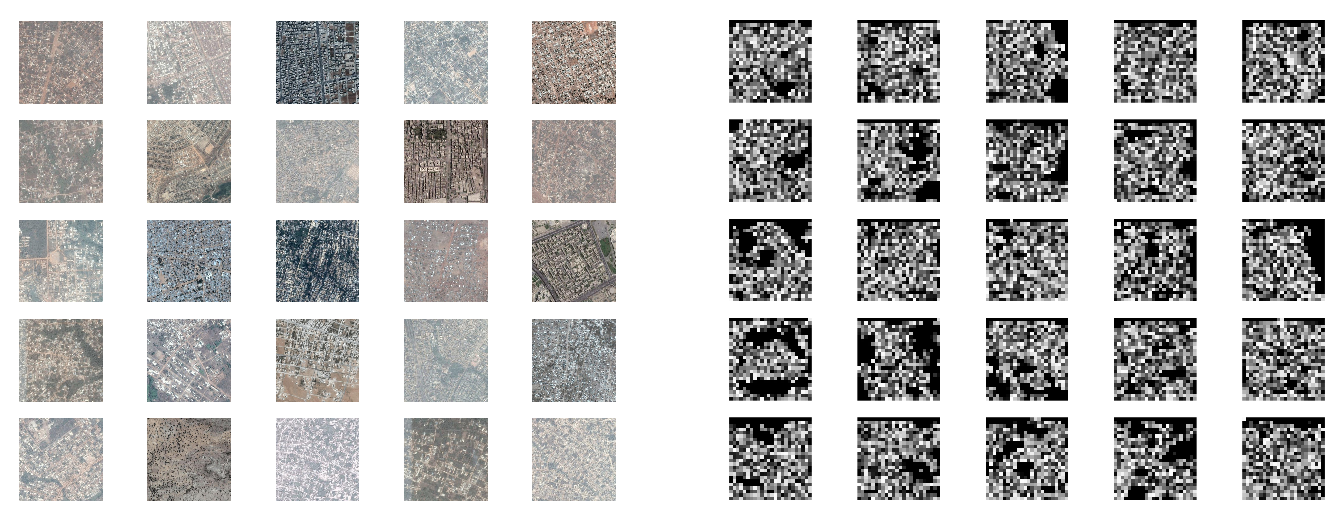}
  \caption{A set of 25 maximally activating images and their corresponding activation maps for a filter in the fifth convolutional layer of the network trained on the 3-class nighttime light intensity prediction task. This filter seems to activate for urban areas, which indicate economic development. }
  \label{fig:filters1}
\end{figure*}

\begin{figure*}[t]
\centering
  \includegraphics[width=\linewidth]{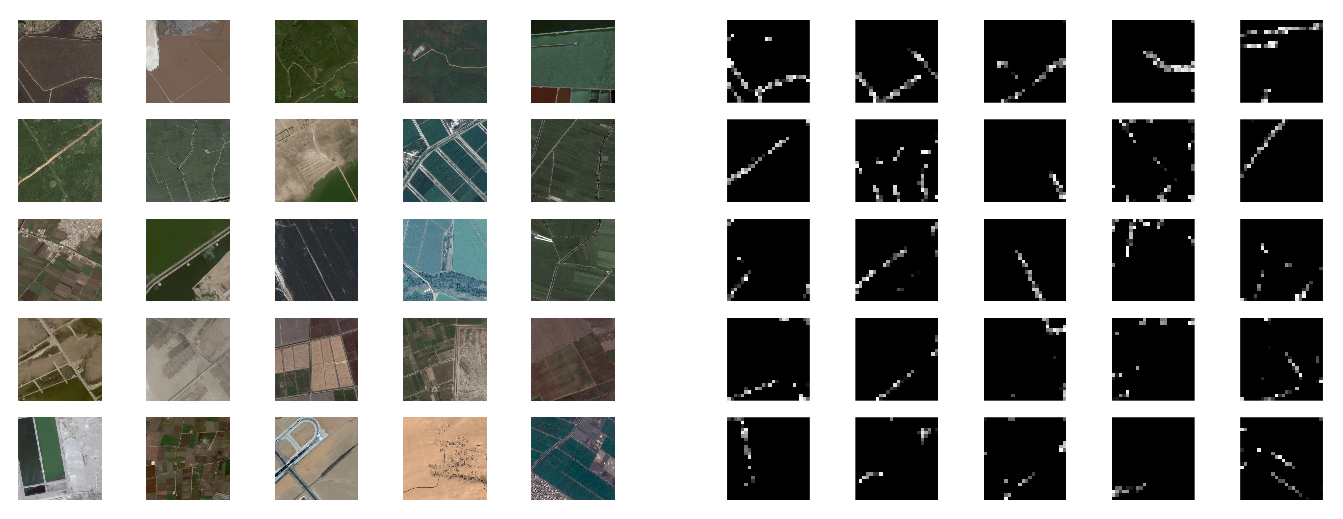}
  \caption{A set of 25 maximally activating images and their corresponding activation maps for a filter in the fifth convolutional layer of the network trained on the 3-class nighttime light intensity prediction task. This filter seems to activate for roads, which are indicative of infrastructure and economic development.}
  \label{fig:filters2}
\end{figure*}

\begin{figure*}[b]
\centering
  \includegraphics[width=\linewidth]{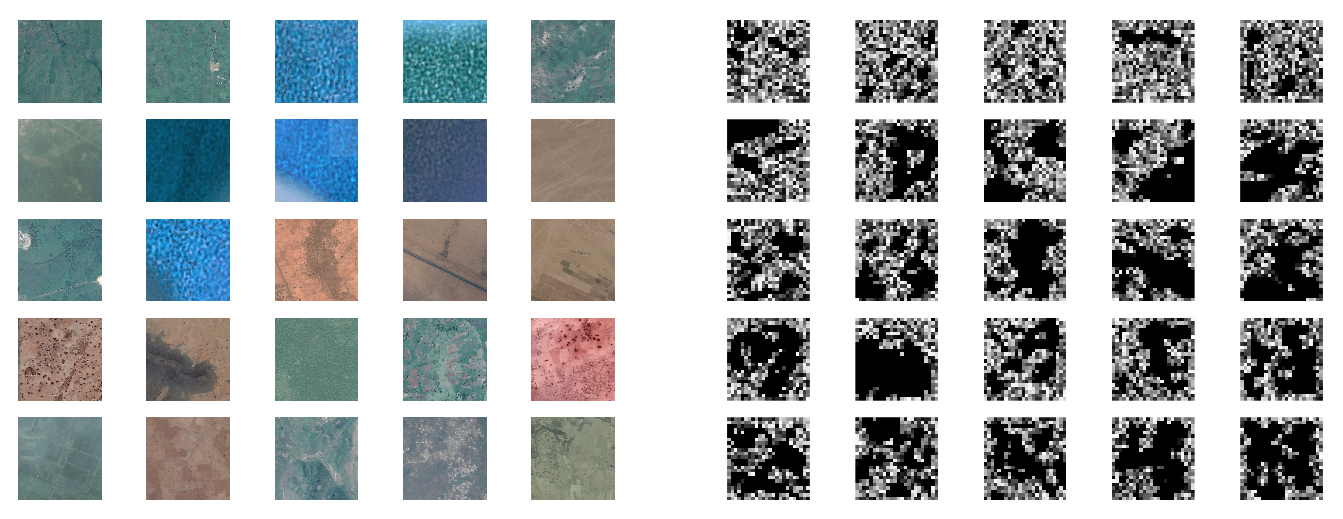}
  \caption{A set of 25 maximally activating images and their corresponding activation maps for a filter in the fifth convolutional layer of the network trained on the 3-class nighttime light intensity prediction task. This filter seems to activate for water, barren, and forested lands, which this filter seems to group together as contributing similarly to nighttime light intensity.}
  \label{fig:filters3}
\end{figure*}

\begin{figure*}[t]
\centering
  \includegraphics[width=\linewidth]{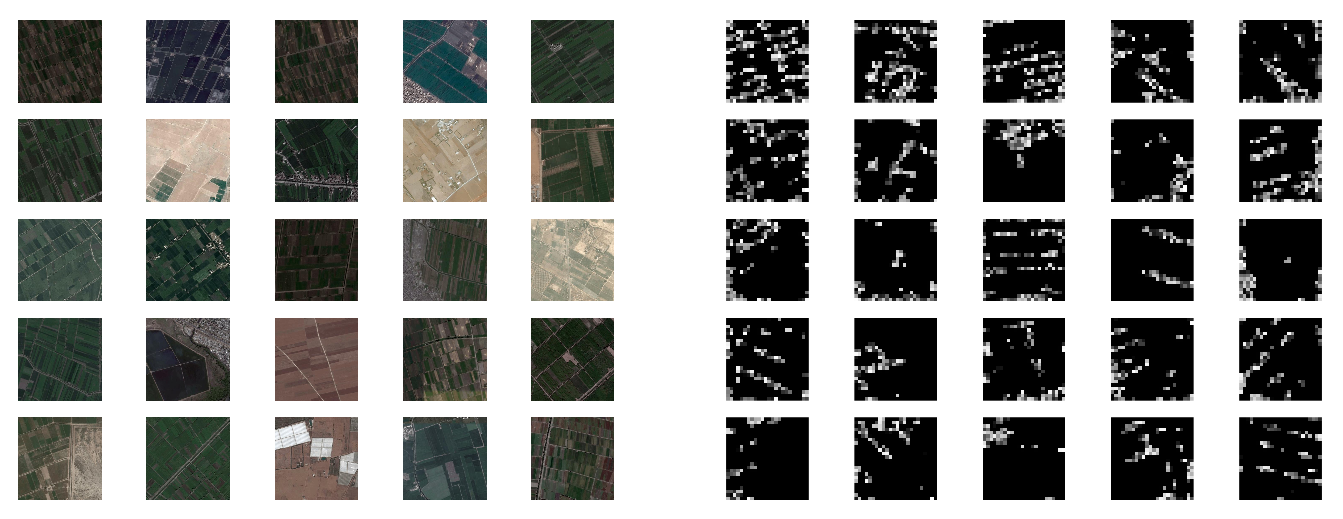}
  \caption{A set of 25 maximally activating images and their corresponding activation maps for a filter in the fifth convolutional layer of the network trained on the 3-class nighttime light intensity prediction task. This filter seems to activate for farmland and for grid-like patterns, which are common in human-made structures.}
  \label{fig:filters4}
\end{figure*}

We provide 25 maximally activating images in the validation set and their activation maps for four filters in our CNN model (Figures \ref{fig:filters1},\ref{fig:filters2},\ref{fig:filters3},\ref{fig:filters4}). The activation maps indicate the locations where the filter activated the most. These filters seem to activate to different terrain types, man-made structures, and roads, all of which can be useful socioeconomic indicators.
\end{document}